\documentclass{article}

     \PassOptionsToPackage{numbers, compress}{natbib}

     \usepackage[preprint]{neurips_2019}

\usepackage[utf8]{inputenc} 
\usepackage[T1]{fontenc}    
\usepackage{hyperref}       
\usepackage{url}            
\usepackage{booktabs}       
\usepackage{amsfonts}       
\usepackage{nicefrac}       
\usepackage{microtype}      
\usepackage{graphicx}
\usepackage{subfigure}
\usepackage{amsthm}

\title{Kernel Wasserstein  Distance}

\author{%
  Jung Hun Oh\textsuperscript{1}\thanks{Corresponding author: Jung Hun Oh  (ohj@mskcc.org). \textsuperscript{$\dagger$} These authors share senior authorship.}, Maryam Pouryahya\textsuperscript{1}, Aditi Iyer\textsuperscript{1}, Aditya P. Apte\textsuperscript{1},\\
  {\bf Allen Tannenbaum\textsuperscript{2$\dagger$}, Joseph O. Deasy\textsuperscript{1$\dagger$}}\\ \textsuperscript{1}Department of Medical Physics,
  Memorial Sloan Kettering Cancer Center, USA \\
  \textsuperscript{2}Departments of Applied Mathematics and Computer Science, Stony Brook University, USA
}

\begin{document}
\setcitestyle{square}
\maketitle

\begin{abstract}
The Wasserstein distance is a powerful metric based on the theory of optimal transport. It gives a natural measure of the distance between two distributions with a wide range of applications. In contrast to a number of the common divergences on distributions such as Kullback-Leibler or Jensen-Shannon, it is (weakly) continuous, and thus ideal for analyzing corrupted data. To date, however, no kernel methods for dealing with nonlinear data have been  proposed via the Wasserstein distance.
In this work, we develop a novel method to compute the \textit{L}\textsuperscript{2}-Wasserstein distance in a kernel space implemented using the {\it kernel trick}. The latter is a general method in machine learning employed to handle data in a nonlinear manner. We evaluate the proposed approach in identifying computerized tomography (CT) slices with dental artifacts in head and neck cancer, performing unsupervised hierarchical clustering on the resulting Wasserstein distance matrix that is computed on imaging texture features extracted from each CT slice.
Our experiments show that the kernel approach  outperforms classical non-kernel approaches in identifying  CT slices with artifacts.
\end{abstract}

\section*{1 Introduction}
Optimal mass transport (OMT) theory is an old research area with its roots in civil engineering (Monge 1781) and economics (Kantorovich 1942) \cite{Peyre2019}. Recently there has been an ever increasing growth in OMT research both theoretically and practically, with impact on numerous fields including medical imaging analysis, statistical physics, machine learning, and genomics \cite{Chen2017,Chen2019,Luise2018,Zhao2013}. The classical OMT problem formulated by  Monge in 1781  concerns finding the optimal way via the minimization of a transportation cost required to move a pile of soil from one site to another \cite{Evans1999,Villani2003,Kantorovich2006, Pouryahya2019}. Let $X$ and $Y$ denote two probability spaces with measures $\mu$ and $\nu$, respectively, and let $c(x,y)$ denote the transportation cost for moving one unit of mass from  $x \in X$ to $y \in Y$. Then the OMT problem seeks to find a (measurable) transport map $T:X\rightarrow Y$ that minimizes the  total transportation
cost $\int_{X} c(x,T(x)){\rm\mu}(dx)$. In 1942, Kantorovich proposed a relaxed formulation that transforms the Monge's nonlinear problem to a linear programming problem \cite{Kantorovich2006}. Based on the Kantorovich's formulation,  the ${\it L^p}$- Wasserstein distance between $\mu$ and $\nu$ on  $\mathcal{R}^{d}$ is defined as:

\begin{equation}
W_{p}^{p} \left(  {\bf\mu} , {\bf\nu} \right) =
 {\rm \inf}_{ \pi \in \Pi (\mu,\nu)}
\int_{{\rm R}^{d} \times {\rm R}^{d}} {\Vert{ x} - { y} \Vert}^{p} d \pi(x,y),
\end{equation}
where $\Pi (\mu,\nu)$ is the set of all joint probability measures $\pi$ on $X \times Y$ whose marginals are $\mu$ and $\nu$.
In particular, in this study, we focus on $\it L^{\rm 2}$- Wasserstein distance in which the squared
Euclidean distance $c(x,y)={\Vert{ x} - { y} \Vert}^{2}$ is the cost function \cite{Mallasto2017}.
Before introducing our proposed kernel Wasserstein distance, we first review some background on the kernel  method.

The \emph{kernel space method} is based on the following idea. Suppose that we are given 
a data set of $n$ samples 
in a native space, denoted by  ${\bf X}=[{\bf x}_1, {\bf x}_2,\cdots, {\bf x}_{\it n}]\in \mathcal{R}^d$. The input data can be mapped (transformed) into a higher dimensional feature space (called the \emph{kernel space}) via a nonlinear mapping function $\phi$ \cite{BAUDAT2000,Oh20091}. Let ${\bf \Phi}$ be  ${\bf \Phi}_{l \times
n}=[\phi({\bf x}_1),\phi({\bf x}_2),\cdots,\phi(\bf x_{\it n})]$, i.e., the transformed data, where $l$ is the number of features in the feature (kernel) space  with $l > d$. To avoid complex data handling in the feature space, and to avoid the explicit computation of the mapping funciton $\phi$, typically one applies the \emph{kernel trick}. More precisely, given any positive definite kernel function $k$, 
one can find an associated mapping function $\phi$  such that 
$k({\bf x},{\bf y})=<\phi{({\bf x})},\phi{({\bf y})}>$ with ${\bf x}, {\bf y}\in 
\mathcal{R}^{d}$ \cite{Scholkopf2000, Rahimi2007}. 
The resulting kernel Gram matrix ${\bf K}$ is defined as:
\begin{equation}
{\bf K} = {\bf \Phi}^{\rm T}{\bf\Phi},\label{Eq:K}
\end{equation}
where the $ij$th element, $\phi({\bf x}_i)^{\rm T}\phi({\bf x}_j)$, is computed using a kernel function $k({{\bf x}_i,{\bf x}_j})= <\phi{({\bf x}_i)},\phi{({\bf x}_j)}>$. Common choices of kernel functions are the polynomial and
radial basis function (RBF) kernels. In this
study, the following RBF kernel will be employed:
\begin{equation}
k({{\bf x}_i,{\bf x}_j})={\rm exp}\big(-\gamma||{{\bf x}_i-{\bf x}_j}||^2 \big),
\end{equation}
where $\gamma > 0$  controls the kernel width. We will fix $\gamma = 1$ in what follows. The mean and the covariance matrix in the feature space
are estimated as:
\begin{equation}
{\bf \mu}=\frac{1}{n}\sum_{i=1}^{n}\phi({\bf x}_{\it i})={\bf\Phi s},
\label{Eq:meansigma}
{~~~\rm and~~~}{\bf\Sigma}=\frac{1}{n}\sum_{i=1}^{n}(\phi({\bf x}_{\it
i})-{\bf \mu})(\phi({\bf x}_{\it i})-{\bf \mu})^{\rm T}={\bf \Phi JJ^{\rm T}\Phi^{\rm T}},
\end{equation}
where ${\bf s}_{n \times 1}=\frac{1}{n}{\bf \vec{1}}^{\rm T}$, $\bf
{J}=\frac{\rm 1}{\sqrt{\it {n}}}(\bf I_{\it n}-s{\bf \vec{1}})$, and
${\bf \vec{1}}=[1,1,\cdots,1]$. Then in Eq.~(\ref{Eq:meansigma}), denoting $\bf {\Phi J}$ by ${\bf W}$ we have
\begin{equation}
\bf {W=\Phi J} = \frac{\rm 1}{\sqrt{\it {n}}}[(\phi({\bf x}_{\rm 1})-{\bf\mu}),
\cdots, (\phi({\bf x}_{\it n})-{\bf\mu})].\label{Eq:W}
\end{equation}
Of note, these equations are used to compute the kernel Wasserstein distance as described in the following section.

\section*{2 Methods}
In this section, we introduce the classical \textit{L}\textsuperscript{2}-Wasserstein distance between Gaussian
measures and then propose a novel approach to compute \textit{L}\textsuperscript{2}-Wasserstein distance in the kernel space (denoted as kernel \textit{L}\textsuperscript{2}-Wasserstein distance). For comparison, we also provide a brief review of the Kullback-Leibler distance in the kernel space that we proposed in \cite{Oh20092}. The code was implemented using MATLAB R2018b.

\subsubsection*{2.1 \textit{L}\textsuperscript{2}-Wasserstein Distance}

For two Gaussian measures,  $\nu _{1} =N_{1} ( {\bf m} _{\rm 1},  {\bf C}  _{\rm 1})$  and  $\nu _{2} = N_{2} (  {\bf m} _{\rm 2},  {\bf C}  _{\rm 2})$ on $ \mathcal{R}^d, $
the \textit{L}\textsuperscript{2}-Wasserstein distance between the two distributions  may be computed as follows \cite{Masarotto2018}:
\begin{eqnarray}
 W_{2} ( \nu _{1},  \nu _{2} ) ^{2}
  =   \Vert{\bf m}_{\rm 1}- {\bf m} _{\rm 2} \Vert  ^{2}+{\rm tr} ({\bf C}  _{\rm1}+  {\bf C}  _{\rm2}-2 (   {\bf C}  _{\rm1}^{\frac{1}{2}}  {\bf C}  _{\rm 2}  {\bf C}  _{\rm1}^{\frac{1}{2}} ) ^{\frac{1}{2}} ),\label{Eq:w2}
 \end{eqnarray}
 where ${\rm tr}$ is the trace. 
 The term ${\rm tr} ({\bf C}  _{\rm1}+  {\bf C}  _{\rm2}-2 (   {\bf C}  _{\rm1}^{\frac{1}{2}}  {\bf C}  _{\rm 2}  {\bf C}  _{\rm1}^{\frac{1}{2}} ) ^{\frac{1}{2}} )$  can be expressed as follows \cite{Dowson1982,Olkin1982,Malago2018}:
\begin{eqnarray}
 {\rm tr} (   {\bf C}  _{1}+  {\bf C}  _{2}-2 (   {\bf C}  _{2}  {\bf C}  _{1} ) ^{\frac{1}{2}} ).  \label{Eq:eq8}
\end{eqnarray}
For convenience, we sketch the proof of this fact.
\begin{proof} By the property of trace,
${\rm tr} (   {\bf C}  _{1}+  {\bf C}  _{2}-2 (   {\bf C}  _{2}  {\bf C}  _{1} )  ^{\frac{1}{2}}) ={\rm tr} (   {\bf C}  _{1} ) +{\rm tr} (   {\bf C}  _{2} ) -2{\rm tr} (   {\bf C}  _{2}  {\bf C}  _{1} ) ^{\frac{1}{2}}.$ Therefore, we need to prove that ${\rm tr} (   {\bf C}  _{2}  {\bf C}  _{1} )^{\frac{1}{2}}={\rm tr}({\bf C}  _{\rm1}^{\frac{1}{2}}  {\bf C}  _{\rm 2}  {\bf C}  _{\rm1}^{\frac{1}{2}} ) ^{\frac{1}{2}}$. Note that ${\bf C}  _{\rm 1}$ and ${\bf C}  _{\rm 2}$ are symmetric positive semidefinite, and ${\bf C}  _{\rm 2}{\bf C}  _{\rm 1}$ is diagonalizable and has nonnegative eigenvalues \cite{Hong1991}.
The eigenvalue decomposition of ${\bf C}  _{2}  {\bf C}  _{1}$  can be computed as
${\bf C}  _{2}  {\bf C}  _{1}{\bf P}={\bf P \Lambda}$ where ${\bf P}$ and $\Lambda$ are the estimated eigenvector and eigenvalue matrices, respectively. Multiplying both sides by ${\bf C}  _{1}^{\frac{1}{2}}$, we have
${\bf C}  _{1}^{\frac{1}{2}}  {\bf C}  _{2}  {\bf C}  _{1}^{\frac{1}{2}}  {\bf C}  _{1}^{\frac{1}{2}}{\bf P}=  {\bf C}  _{1}^{\frac{1}{2}} {\bf P \Lambda}.$ That is, ${\bf C}  _{1}^{\frac{1}{2}}  {\bf C}  _{2}  {\bf C}  _{1}^{\frac{1}{2}} $ has an eigenvector matrix ${\bf C}  _{1}^{\frac{1}{2}}{\bf P}$ and an eigenvalue matrix ${\bf \Lambda}$ which is the same as that of ${\bf C}  _{2}  {\bf C}  _{1}$ \cite{Gosson2008}. Let $\lambda_{1}, \lambda_{2}, ..., \lambda_{k}$ be the distinct eigenvalues of  ${\bf C}  _{2}  {\bf C}  _{1}$. Then the eigenvalues of $({\bf C}  _{2}  {\bf C}  _{1} )^{\frac{1}{2}}$  are $\sqrt{\lambda_{1}}, \sqrt{\lambda_{2}}, ...,\sqrt{\lambda_{k}}$. Therefore, ${\rm tr} (   {\bf C}  _{2}  {\bf C}  _{1} )^{\frac{1}{2}}={\rm tr}({\bf C}  _{\rm1}^{\frac{1}{2}}  {\bf C}  _{\rm 2}  {\bf C}  _{\rm1}^{\frac{1}{2}} ) ^{\frac{1}{2}}$.
\end{proof}
In particular, when ${\bf C}  _{1}={\bf C}  _{2}$, we have $W_{2} ( \nu _{1},  \nu _{2} )^{2}
  =   \Vert{\bf m}_{\rm 1}- {\bf m} _{\rm 2} \Vert^{2}.$

\subsubsection*{2.2 \textit{L}\textsuperscript{2}-Wasserstein Distance in Kernel Space}

Suppose that we are given two Gaussian measures, $k\nu_{1}$ and $k\nu_{2} \in  \mathcal{R}^{\it
l}$, in the kernel space with mean ${\bf \mu}_{\it i}$ and covariance matrix ${\bf \Sigma}_{\it i}$, for ${\it i}$=1 and 2,
and two sets of sample data  in the native space, ${\bf X}=[{\bf x}_1, {\bf x}_2,\cdots, {\bf x}_{\it n}], \; {\bf Y}=[{\bf y}_1, {\bf y}_2,\cdots, {\bf y}_{\it m}] \in \mathcal{R}^d$ associated with $k\nu_{1}$ and $k\nu_{2}$, respectively.
Then, as in Eq.~(\ref{Eq:w2}), the \textit{L}\textsuperscript{2}-Wasserstein distance between the two distributions is given by:
\begin{eqnarray}
 W_{2} ( k\nu _{1},  k\nu _{2} ) ^{2}
  =   \Vert{\bf \mu}_{\rm 1}- {\bf \mu} _{\rm 2} \Vert  ^{2}+{\rm tr} ({\bf \Sigma}  _{\rm1}+  {\bf \Sigma}  _{\rm2}-2 (   {\bf \Sigma}  _{\rm1}^{\frac{1}{2}}  {\bf \Sigma}  _{\rm 2}  {\bf \Sigma}  _{\rm1}^{\frac{1}{2}} ) ^{\frac{1}{2}} ),\label{Eq:kw2}
 \end{eqnarray}
 where the definitions of ${\bf \mu} _{\it i}$ and ${\bf \Sigma} _{\it i}$ are shown in Eq.~(\ref{Eq:meansigma}). Note that ${\rm tr}({\bf \Sigma}  _{\rm1}^{\frac{1}{2}}  {\bf \Sigma}  _{\rm 2}  {\bf \Sigma}  _{\rm1}^{\frac{1}{2}} ) ^{\frac{1}{2}}={\rm tr}\left(   {\bf\Sigma}  _{2}  {\bf\Sigma}  _{1} \right) ^{\frac{1}{2}}$. The first term, $\Vert{\bf\mu} _{\rm 1}- {\bf\mu} _{\rm 2} \Vert  ^{2}$, in Eq.~(\ref{Eq:kw2}) can be obtained as follows \cite{Gretton2012}:
\begin{eqnarray}
\Vert  {\bf\mu} _{1}- {\bf\mu} _{2} \Vert  ^{2}= \Vert  {\bf\mu} _{1} \Vert  ^{2}-2 {\bf\mu} _{1}^{\rm T} {\bf\mu} _{2}+ \Vert  {\bf\mu} _{2} \Vert  ^{2}.\label{Eq:mmd}
\end{eqnarray} Via a simple computation, Eq.~(\ref{Eq:mmd}) can be expressed as:
\begin{eqnarray}
 \Vert  {\bf\mu} _{1}- {\bf\mu} _{2} \Vert  ^{2}=\frac{1}{n^{2}} \sum _{i=1}^{n} \sum _{j=1}^{n}{\it k} \left( {\bf x}_{i},{\bf x}_{j} \right) -\frac{2}{nm} \sum _{i=1}^{n} \sum _{j=1}^{m}{\it k} \left( {\bf x}_{i},{\bf y}_{j} \right) +\frac{1}{m^{2}} \sum _{i=1}^{m} \sum _{j=1}^{m}{\it k} \left( {\bf y}_{i},{\bf y}_{j} \right).  \label{Eq:firstterm}
 \end{eqnarray}

Now we simplify the last term in Eq.~(\ref{Eq:kw2}) by using  Eq.~(\ref{Eq:meansigma}) :\\
$~~~~~~~~~~~~~~~~~~~~
{\rm tr} (   {\bf\Sigma}  _{1}+  {\bf\Sigma}  _{2}-2 (   {\bf\Sigma}  _{2}  {\bf\Sigma}  _{1} ) ^{\frac{1}{2}})
={\rm tr} \left(   {\bf\Sigma}  _{1} \right) +{\rm tr} \left(   {\bf\Sigma}  _{2} \right) -2 {\rm tr} \left(   {\bf\Sigma}  _{2}  {\bf\Sigma}  _{1} \right) ^{\frac{1}{2}}
$
\begin{eqnarray}
\nonumber
~~~~~~~~~~~~~~~~~~&=& {\rm tr} \left(  {\bf\Phi} _{1}{\bf J}_{1}{\bf J}_{1}^{\rm T} {\bf\Phi} _{1}^{\rm T} \right) +{\rm tr} \left(  {\bf\Phi} _{2}{\bf J}_{2}{\bf J}_{2}^{\rm T} {\bf \Phi} _{2}^{\rm T} \right) -2 {\rm tr} \left(  {\bf\Phi} _{2}{\bf J}_{2}{\bf J}_{2}^{\rm T} {\bf \Phi} _{2}^{\rm T} {\bf \Phi} _{1}{\bf J}_{1}{\bf J}_{1}^{\rm T} {\bf\Phi} _{1}^{\rm T} \right) ^{\frac{1}{2}} \\
~~~~~~~~~~~~~~~~~~&=& {\rm tr} \left( {\bf J}_{1}{\bf J}_{1}^{\rm T} {\bf\Phi} _{1}^{\rm T} {\bf\Phi} _{1} \right) +{\rm tr} \left( {\bf J}_{2}{\bf J}_{2}^{\rm T} {\bf\Phi} _{2}^{\rm T} {\bf\Phi} _{2} \right) -2 {\rm tr} \left(  {\bf\Phi} _{2}{\bf J}_{2}{\bf J}_{2}^{\rm T}{\bf K}_{21}{\bf J}_{1}{\bf J}_{1}^{\rm T} {\bf\Phi} _{1}^{\rm T} \right) ^{\frac{1}{2}} \\ \nonumber
~~~~~~~~~~~~~~~~~~&=& {\rm tr} \left( {\bf J}_{1}{\bf J}_{1}^{\rm T}{\bf K}_{11} \right) +{\rm tr} \left( {\bf J}_{2}{\bf J}_{2}^{\rm T}{{\bf K}_{22}} \right) -2 {\rm tr} \left(  {\bf\Phi} _{2}{\bf R} {\bf \Phi} _{1}^{\rm T} \right) ^{\frac{1}{2}},
\end{eqnarray}
where
${\bf R}={\bf J}_{2}{\bf J}_{2}^{\rm T}{\bf K}_{21}{\bf J}_{1}{\bf J}_{1}^{\rm T}$ and ${\bf K} _{\it ij}= {\bf\Phi} _{i}^{\rm T}{\bf\Phi_{\it j}}$. Note that  ${\bf\Sigma_{\rm 1}}$ and ${\bf\Sigma_{\rm 2}}$ are symmetric positive semidefinite. Therefore, ${\bf\Sigma_{\rm 2}}{\bf\Sigma_{\rm 1}}={\bf\Phi} _{2}{\bf R} {\bf\Phi} _{1}^{\rm T}$ is diagonalizable and has nonnegative eigenvalues.
Suppose that  the eigenvector and eigenvalue matrices of ${\bf \Phi} _{2}{\bf R} {\bf\Phi} _{1}^{\rm T}$ are $\bf 	\tilde{P}$ and ${\bf 	\tilde{\Lambda}}$:
\begin{eqnarray}
{\bf \Phi} _{2}{\bf R} {\bf \Phi} _{1}^{\rm T}{\bf 	\tilde{P}}={\bf 	\tilde{P} 	\tilde{\Lambda}}.
\end{eqnarray}
By multiplying both sides by ${\bf \Phi} _{1}^{\rm T}$, we have ${\bf \Phi} _{1}^{\rm T} {\bf \Phi} _{2}{\bf R} {\bf \Phi} _{1}^{\rm T}{\bf 	\tilde{P}}= {\bf \Phi} _{1}^{\rm T}{\bf 	\tilde{P} 	\tilde{\Lambda}}$. That is, the eigenvalue matrix of ${\bf \Phi} _{1}^{\rm T} {\bf \Phi} _{2}{\bf R}$ is the same as that of ${\bf \Phi} _{2}{\bf R} {\bf \Phi} _{1}^{\rm T}$ with ${\bf 	\tilde{\Lambda}}$. Therefore, the following equations hold:
\begin{eqnarray}{\rm tr} \left(  {\bf \Phi} _{2}{\bf R} {\bf \Phi} _{1}^{\rm T} \right) ^{\frac{1}{2}}={\rm tr} \left(  {\bf \Phi} _{1}^{\rm T} {\bf \Phi} _{2}{\bf R} \right) ^{\frac{1}{2}}={\rm tr} \left( {\bf K}_{12}{\bf R} \right) ^{\frac{1}{2}}={\rm tr} \left( {\bf K}_{12}{\bf J}_{2}{\bf J}_{2}^{\rm T}{\bf K}_{21}{\bf J}_{1}{\bf J}_{1}^{\rm T} \right) ^{\frac{1}{2}}. \label{Eq:secondterm}
\end{eqnarray}

Finally, using  Eq.~(\ref{Eq:firstterm}) and Eq.~(\ref{Eq:secondterm}), the kernel $\textit{L}\textsuperscript{2}$-Wasserstein distance can be expressed as:  \\
$ W_{2} ( k\nu _{1},  k\nu _{2} ) ^{2}$=$   \Vert  {\bf \mu} _{1}- {\bf \mu} _{2} \Vert  ^{2}+{\rm tr} (   {\bf\Sigma}  _{1}+  {\bf\Sigma}  _{2}-2 (   {\bf\Sigma}  _{1}^{\frac{1}{2}}  {\bf\Sigma}  _{2}  {\bf\Sigma}  _{1}^{\frac{1}{2}} ) ^{\frac{1}{2}}) $
\begin{eqnarray}
{~~~~~~~~~~~~~~~~~~~~}=\frac{1}{n^{2}} \sum _{i=1}^{n} \sum _{j=1}^{n}{\it k} \left( {\bf x}_{i},{\bf x}_{j} \right) -\frac{2}{nm} \sum _{i=1}^{n} \sum _{j=1}^{m}{\it k} \left( {\bf x}_{i},{\bf y}_{j} \right) +\frac{1}{m^{2}} \sum _{i=1}^{m} \sum _{j=1}^{m}{\it k} \left( {\bf y}_{i},{\bf y}_{j} \right)+ \\{\rm tr} \left( {\bf J}_{1}{\bf J}_{1}^{\rm T}{\bf K}_{11} \right) +{\rm tr} \left( {\bf J}_{2}{\bf  J}_{2}^{\rm T}{\bf K}_{22} \right) -2{\rm tr} \left( {\bf K}_{12}{\bf J}_{2}{\bf J}_{2}^{\rm T}{\bf K}_{21}{\bf J}_{1}{\bf J}_{1}^{\rm T} \right) ^{\frac{1}{2}}.  \nonumber \label{Eq:final_kwd}
\end{eqnarray}
In a special case, when ${\bf \Sigma}  _{1}={\bf \Sigma}  _{2}$, we have $W_{2} ( k\nu _{1},  k\nu _{2} )^{2}
  =   \frac{1}{n^{2}} \sum _{i=1}^{n} \sum _{j=1}^{n}{\it k} \left( {\bf x}_{i},{\bf x}_{j} \right) -\frac{2}{nm} \sum _{i=1}^{n} \sum _{j=1}^{m}{\it k} \left( {\bf x}_{i},{\bf y}_{j} \right) +\frac{1}{m^{2}} \sum _{i=1}^{m} \sum _{j=1}^{m}{\it k} \left( {\bf y}_{i},{\bf y}_{j} \right) $.

\subsection*{2.3 Kullback-Leibler  Divergence in Kernel Space}

Kullback-Leibler (KL) divergence is another type of method to compare two probability distributions \cite{Bauckhage2013}. It is not a distance measure due to its asymmetric nature. 
Let $P$ and $Q$ be   two continuous probability distributions with the corresponding probability densities   $p(x)$ and $q(x)$, respectively. Then  the \emph{KL divergence or relative 
entropy}  of $P$ and
$Q$ over the same
variable $x$ is  defined as:
\begin{equation}
D_{\mathrm{KL}}(P\|Q) = \int_{-\infty}^{\infty} p({ x}) \log \frac{p({
x})}{q({x})}{ dx},
\end{equation}
where $D_{\mathrm{KL}}(P\|Q)$ equals zero if and only if $P=Q$.
Given two Gaussian measures in the kernel space, ${N}_1({\bf\mu}_1,{\bf\Sigma}_1)$ and
${N}_2({\bf\mu}_2,{\bf\Sigma}_2)$ with both  $\in \mathcal{R}^{\it
l}$, the KL divergence may be computed to be:
\begin{eqnarray}
D_{\rm KL}({N}_1||{N}_2)=\frac{1}{2}\big\{({\bf\mu}_1-{\bf\mu}_2)^{\bf
{\rm T}}{\bf\Sigma}_2^{-1}({\bf\mu}_1-{\bf\mu}_2)+{\rm log}\frac{|{\bf\Sigma}_2|}{|{\bf\Sigma}_1|}
+{\rm tr}[{\bf\Sigma}_1{\bf\Sigma}_2^{-1}]-l\big\},\label{Eq:kl}
\end{eqnarray}
where $|{\bf \Sigma_{\it i}}|$ is the determinant
of covariance matrix ${\bf\Sigma}_{\it i}$. Note that $l$ is the number of features in the kernel space and indeed this is unknown. Importantly, this variable is canceled out when Eq.~(\ref{Eq:kl}) is completely solved, which  will be  explained later. In a special case, when ${\bf\Sigma}={\bf\Sigma}_1={\bf\Sigma}_2, D_{\rm KL}({N}_1||{N}_2)=\frac{1}{2}\big\{({\bf\mu}_1-{\bf\mu}_2)^{\bf
{\rm T}}{\bf\Sigma}^{-1}({\bf\mu}_1-{\bf\mu}_2)\}$.


\subsection*{\bf Singularity Problem } 

In many  real problems, the number of samples is considerably smaller than the number of features, leading to the covariance matrix being singular, and therefore non-invertible. Typically, one deals with data in a higher dimensional space as in this study. To avoid the singularity problem, several methods have been proposed \cite{Ye20042, Li2005}. 
In this work, we employ a simple method by adding some positive values
 to the diagonal elements of the covariance
matrix \cite{Ye20041}. Therefore, the modified covariance matrix is of
full rank and invertible. Set
\begin{eqnarray}\label{Eq:H}
\bf {H}=\bf \Sigma+\rho{\bf I}_{\it l} =\bf {\Phi JJ^{\rm T}\Phi^{\rm T}}+\rho{\bf I}_{\it l}
= {\bf WW^{\rm T}}+\rho{\bf I}_{\it l}= {\bf \Phi S
\Phi^{\rm T}}+\rho{\bf I}_{\it l},
\end{eqnarray}
where $\bf W=\Phi J$ as in Eq.~(\ref{Eq:W}), $\bf S=JJ^{\rm T}$, $l$ is the number of features, and ${\bf I}_{\it l}$ is an $l \times l$ identity
matrix. In this study, $\rho =0.1$ is used. 

In the next computation, we will employ the {\it Woodbury formula} \cite{Lai1997}, which we now review. Let 
 $\bf A$ be a 
square $r \times r$ invertible matrix, and let $\bf U$ and $\bf V$ 
be matrices of size $r \times k$ with $k \leq r$. Assume that the $k \times
k$ matrix ${\bf \Sigma}={\bf I}_k+\beta {\bf V^{\rm T}A^{\rm -1}U}$ is invertible, where $\beta$ is an
arbitrary scalar. Then the {\it Woodbury formula} states that
\begin{eqnarray}\nonumber
({\bf A}+\beta {\bf UV^{\rm T})}^{\rm -1}={\bf A^{\rm -1}}-\beta {\bf
A^{\rm -1}U\Sigma^{\rm -1}V^{\rm T}A^{\rm -1}}.
\end{eqnarray}

Accordingly, utilizing this formula, we can compute the inverse of
$\bf H$:
\begin{eqnarray}\label{Eq:invH}\bf {H^{\rm -1}}&=& (\rho{\bf I}_l+\bf {\Phi
JJ^{\rm T}\Phi^{\rm T}})^{\rm -1}\\\nonumber &=& (\rho{\bf I}_l+{\bf
WW^{\rm T}})^{-1}~~~ \lhd{\rm apply~}Woodbury~formula\\\nonumber
&=& (\rho{\bf I}_l)^{-1}-(\rho{\bf
I}_l)^{-1}{\bf W}({\bf I}_n+{\bf W^{\rm T}}(\rho{\bf I}_l)^{-1}{\bf
W)^{\rm -1}W^{\rm T}}(\rho{\bf I}_l)^{-1}\\\nonumber
&=& \rho^{-1}({\bf I}_l-\rho^{-1}{\bf W}({\bf I}_n+\rho^{-1}{\bf
W^{\rm T}W})^{-1}{\bf W^{\rm T}})\\\nonumber &=& \rho^{-1}({\bf I}_l-{\bf
W}(\rho {\bf I}_n+{\bf W^{\rm T}W})^{-1}{\bf W^{\rm T}})\\\nonumber &=&
\rho^{-1}({\bf I}_l-{\bf \Phi} {\bf JM^{\rm -1}J^{\rm T}\Phi^{\rm T}})\\\nonumber &=&
\rho^{-1}({\bf I}_l-{\bf \Phi} {\bf B} {\bf \Phi}^{{\rm T}}),
\end{eqnarray}
where $\bf B=JM^{\rm -1}J^{\rm T}$, ${\bf M}=\rho {\bf I}_n+{\bf
W^{\rm T}W}=\rho {\bf I}_n+{\bf J^{\rm T}}{\bf \Phi}^{\rm T}{\bf \Phi} {\bf J}=\rho {\bf
I}_n+{\bf J^{\rm T}KJ}$, and $n$ is the number of samples. In ${\bf H^{\rm -1}}$, some mapping functions are still left. These will be replaced with kernel functions when kernel KL divergence is completely solved.

\subsection*{Calculation of Kernel KL Divergence}

In the Experiments section below, we will compare Wasserstein and KL divergence based kernel methods. Accordingly, we sketch the necessary theory for the kernel KL divergence approach.

Suppose that we are given two Gaussian measures in the kernel space, ${N}_1({\bf\mu}_1,{\bf\Sigma}_1), \;
{N}_2({\bf\mu}_2,{\bf\Sigma}_2) \in \mathcal{R}^{\it
l}$, consisting of $n$ and $m$ samples, respectively. Assume that ${\bf\Sigma}_1$ and ${\bf\Sigma}_2$ are singular in the higher dimensional space. Let ${\bf H}_1$ and ${\bf H}_2$ denote the
approximate covariance matrices for the two distributions as in Eq.~(\ref{Eq:H}). Then, the kernel KL divergence  is expressed as follows:
\begin{eqnarray}
2D_{\rm {KL}}({N}_1||{N}_2)=(\mu_1-\mu_2)^{\bf
{\rm T}}{\bf
 H}_2^{-1}(\mu_1-\mu_2)+{\rm log}\frac{|{\bf H}_2|}{|{\bf H}_1|}
+{\rm tr}[{\bf H}_1{\bf H}_2^{-1}]-{\it l}.
\end{eqnarray}
We now solve each term separately: (1) $(\mu_1-\mu_2)^{\bf
{\rm T}}{\bf
 H}_2^{-1}(\mu_1-\mu_2)$, (2) ${\rm log}\frac{|{\bf H}_2|}{|{\bf H}_1|}$, and (3)
${\rm tr}[{\bf H}_1{\bf H}_2^{-1}]$.
The first term consists of
four sub-terms:
\begin{eqnarray}
(\mu_1-\mu_2)^{\bf {\rm T}}{\bf
 H}_2^{-1}(\mu_1-\mu_2)=
 \mu_1^{\rm T}{\bf H}_2^{-1}\mu_1+\mu_2^{\rm T}{\bf H}_2^{-1}\mu_2-\mu_1^{\rm T}{\bf H}_2^{-1}\mu_2-\mu_2^{\rm T}{\bf
 H}_2^{-1}\mu_1.
\end{eqnarray}
Substituting Eq.~(\ref{Eq:meansigma}) and  Eq.~(\ref{Eq:invH}) into
each sub-term $\mu_i^{{\rm T}}{\bf H}_j^{-1}\mu_k$, we have
\begin{eqnarray}
\mu_i^{{\rm T}}{\bf H}_j^{-1}\mu_k&=&{\bf s}_i^{\rm T}{\bf \Phi}_i^{\rm
T}\rho ^{-1}({\bf I}_l-{\bf \Phi}_j {\bf B}_j{\bf \Phi}_j^{\rm T}){\bf \Phi}_k {\bf
s}_k\\\nonumber&=&\rho^{-1}({\bf s}_i^{\rm T}{\bf K}_{ik}{\bf
s}_k-{\bf s}_i^{\rm T}{\bf K}_{ij}{\bf B}_j{\bf K}_{jk}{\bf
s}_k)\\\nonumber&=&\rho^{-1}\theta_{ijk},
\end{eqnarray}
where $\theta_{ijk}={\bf s}_i^{\rm T}{\bf K}_{ik}{\bf
s}_k-{\bf s}_i^{\rm T}{\bf K}_{ij}{\bf B}_j{\bf K}_{jk}{\bf
s}_k$.
As a result, all the mapping functions in the first term
can be replaced with kernel functions.
For the second term, we should compute the determinant of ${\bf H}$. To accomplish this, we use a simple trick by computing the
determinant of ${\bf H}^{-1}$ instead of ${\bf H}$.
\begin{eqnarray}
|{\bf H{\rm _1}}^{-1}|&=&|\rho^{-1}({\bf I}_l-{\bf \Phi_{\rm 1} B_{\rm 1}
\Phi_{\rm 1}^{\rm T}})|\\\nonumber &=&\rho^{-l}|{\bf I}_l-{\bf \Phi_{\rm 1}
B_{\rm 1}\Phi_{\rm 1}^{\rm T}}| ~~~\lhd {\rm by~}|d{\bf A|}={\it d^r}{\rm
|{\bf A}|} {~~\rm for~} {\bf A}_{\it r \times r}\\\nonumber
&=&\rho^{-l}|{\bf I}_l-{\bf Q_{\rm 1} \Phi_{\rm 1}^{\rm T}}|
\\\nonumber&=&\rho^{-l}|{\bf I}_n-{\bf \Phi_{\rm 1}^{\rm T} Q_{\rm 1} }|~~~~~\lhd {\rm by~ }|{\bf I}_k-{\bf AB^{\rm T}}|={\rm }|{\bf I}_r-{\bf B^{\rm T}A}| {\rm ,~} {\bf A} {~\rm and~} {\bf B} {~\rm with ~size~}{\it k \times r}
\\\nonumber&=&\rho^{-l}|{\bf I}_n-{\bf \Phi_{\rm 1}^{\rm T} \Phi_{\rm 1} B_{\rm 1} }|
\\\nonumber&=&\rho^{-l}|{\bf I}_n-{\bf K_{\rm 11}B_{\rm 1} }|,
\nonumber
\end{eqnarray}
where ${\bf Q_{\rm 1}}={\bf \Phi}_{1} {\bf B}_{1}$. Now we compute  $|{\bf H}_{1}|$ as follows:
\begin{equation}
{\bf |H_{\rm 1}|=\frac{\rm 1}{|H{_{\rm 1}}^{\rm -1}|}}=\frac{\rho^l}{|{\bf I}_n-{\bf K_{\rm 11}B_{\rm 1} }|}.
\end{equation}
By taking $\mathrm {logarithm }$ of $|{\bf H}_{1}|$, we have
\begin{equation}
{\rm log |{\bf H_{\rm 1}}|}={\rm log}\frac{\rho^l}{|{\bf I}_n-{\bf K_{\rm 11}B_{\rm 1} }|}=l{\rm
log}\rho-{\rm log}|{\bf I}_n-{\bf K_{\rm 11}B_{\rm 1} }|.
\end{equation}
Therefore, we have the second
term composed of kernel functions:
\begin{eqnarray}
{\rm log}\frac{|{\bf H}_2|}{|{\bf H}_1|}={\rm
log|{\bf H}_2|-log|{\bf H}_1|}= {\rm log}|{\bf I}_{n}-{\bf
K_{\rm 11}B_{\rm 1}}|-{\rm log}|{\bf I}_{\it m}-{\bf K}_{\rm 22}{\bf B}_{\rm 2}|.
\end{eqnarray}
The third term can be replaced with kernel functions using
properties of trace:
\begin{eqnarray}
{\rm tr}[{\bf H}_{\rm 1}{\bf H}_{\rm 2}^{\rm -1}] &=&
{\rm tr}[({\bf\Phi}_1{\bf S}_1{\bf\Phi}_1^{\rm T}+\rho {\bf I}_{\it l})\rho^{\rm -1}({\bf I}_{\it
l}-{\bf \Phi}_{\rm 2}{\bf B}_{\rm 2}{\bf \Phi}_{\rm 2}^{\rm T})]~~~~~~~~~~
\\\nonumber
&=&
\rho^{-1}{\rm tr}[{\bf\Phi}_1{\bf S}_1{\bf \Phi}_1^{\rm T}]-\rho^{-1}{\rm tr}[{\bf \Phi}_1{\bf S}_1{\bf \Phi}_1^{\rm T}{\bf \Phi}_2{\bf B}_2{\bf \Phi}_2^{\rm T}]
+{\it l}-{\rm tr}[{\bf\Phi}_{\rm 2}{\bf B}_{\rm2}{\bf\Phi}_{\rm 2}^{\rm T}]\\\nonumber &=&
\rho^{-1}{\rm tr}[{\bf S}_{\rm 1}{\bf K}_{\rm 11}]-\rho^{-1}{\rm tr}[{\bf S}_{\rm 1}
{\bf K}_{12}{\bf B}_{2}{\bf K}_{\rm 21}]+{\it l}-{\rm tr}[{\bf B}_{\rm 2}{\bf K}_{\rm 22}].
\end{eqnarray}

Consequently, we have solved kernel KL divergence by replacing all mapping functions with kernel functions and it is expressed as:
\\$
2D_{\rm {KL}}({N}_1||{N}_2)=(\mu_1-\mu_2)^{\bf
{\rm T}}{\bf
 H}_2^{-1}(\mu_1-\mu_2)+{\rm log}\frac{|{\bf H}_2|}{|{\bf H}_1|}
+{\rm tr}[{\bf H}_1{\bf H}_2^{-1}]-{\it l}
$
\begin{eqnarray}
{~~~~~~~~~~~~~~~~~~~~~~~~~}=\rho^{-1}(\theta_{121}+\theta_{222}-\theta_{122}-\theta_{221})+
{\rm log}|{\bf I}_{n}-{\bf
K_{\rm 11}B_{\rm 1}}|-{\rm log}|{\bf I}_{\it m}-{\bf K}_{\rm 22}{\bf B}_{\rm 2}|\\\nonumber
+\rho^{-1}{\rm tr}[{\bf S}_{\rm 1}{\bf K}_{\rm 11}]-\rho^{-1}{\rm tr}[{\bf
S}_{\rm 1}{\bf K}_{12}{\bf B}_{2}{\bf K}_{\rm 21}]-{\rm tr}[{\bf B}_{\rm 2}{\bf K}_{\rm 22}].
\end{eqnarray}
Note that the  $l$ is canceled out. Moreover, since $D_{\mathrm{KL}}(P\|Q)	\neq D_{\mathrm{KL}}(Q\|P)$, in this study an average value of two KL measures is used:
$J_{\mathrm{KL}}(P\|Q)= \frac{1}{2}\big\{D_{\mathrm{KL}}(P\|Q)	+ D_{\mathrm{KL}}(Q\|P)\big\}$ \cite{Rakocevic2013}.

\begin{figure}[t]
	\centering
	\includegraphics[width=0.7\textwidth]{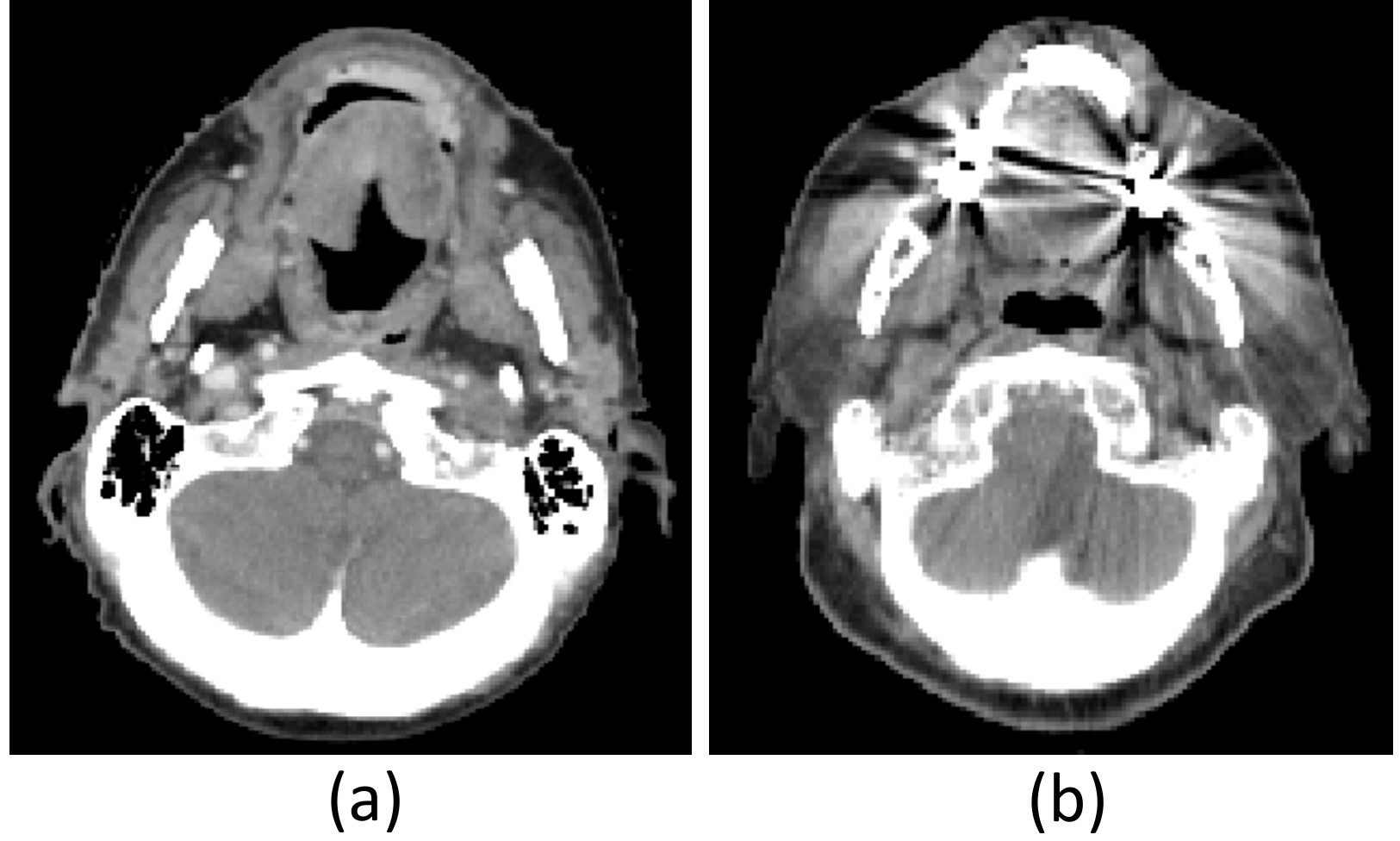}
	\caption{Representative clean (a) and noisy (b) slices.}
	\label{fig:1}
\end{figure}

\section*{3 Experiments}
\subsection*{3.1 Data}
We investigated the utility of kernel \textit{L}\textsuperscript{2}-Wasserstein distance to identify slices with dental artifacts in computerized tomography (CT) scans in head and neck cancer. Serious image degradation caused by metallic fillings or crowns in teeth is a common problem in CT images. We analyzed 1164 axial slices from 44 CT scans that were collected from 44 patients with head and cancer  who were treated in our institution. This retrospective study was approved by the institutional review board and  informed consent was obtained from all patients. Before the analysis, each CT slice was labeled as {\it noisy} or {\it clean} based on the presence of dental artifacts by a medical imaging expert, resulting in 276 noisy and 888 clean slices.
Figure 1 shows representative noisy and clean CT slices from two different scans.

\begin{table}[]
\caption {GLCM-based 25 texture features used in this study.} \label{tab:title}
\begin{tabular}{lllll}
\hline
\textbf{No} & \textbf{Features}   & \textbf{No} & \textbf{Features}                                                                &  \\ \hline
1           & Auto-correlation    & 14          & Inverse Difference Moment                                                        &  \\
2           & Joint Average       & 15          & \begin{tabular}[c]{@{}l@{}}First Informal   Correlation\end{tabular}           &  \\
3           & Cluster Prominence  & 16          & \begin{tabular}[c]{@{}l@{}}Second Informal  Correlation\end{tabular}          &  \\
4           & Cluster Shade       & 17          & \begin{tabular}[c]{@{}l@{}}Inverse Difference Moment   Normalized\end{tabular} &  \\
5           & Cluster Tendency    & 18          & Inverse Difference Normalized                                                    &  \\
6           & Contrast            & 19          & Inverse Variance                                                                 &  \\
7           & Correlation         & 20          & Sum Average                                                                      &  \\
8           & Difference Entropy  & 21          & Sum Entropy                                                                      &  \\
9           & Dissimilarity       & 22          & Sum Variance                                                                     &  \\
10          & Difference Variance & 23          & Haralick Correlation                                                             &  \\
11          & Joint Energy        & 24          & Joint Maximum                                                                    &  \\
12          & Joint Entropy       & 25          & Joint Variance                                                                   &  \\
13          & Inverse Difference  &             &                                                                                  &  \\ \hline
\end{tabular}
\end{table}
\begin{figure}[t]
	\centering
	\includegraphics[width=1.0\textwidth]{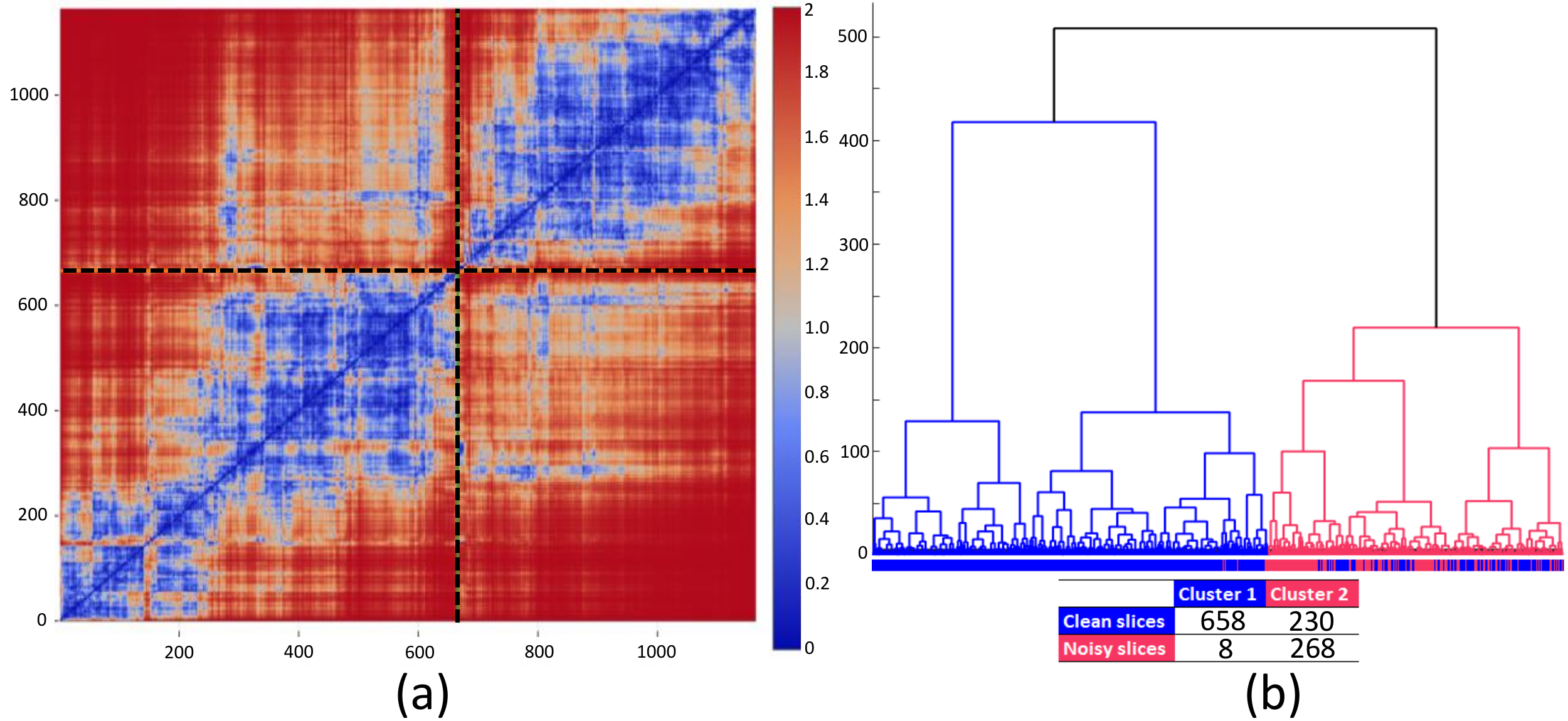}
	\caption{Results for kernel Wasserstein distance: (a)  heatmap for the resulting distance matrix  and (b) hierarchical clustering result conducted using the  distance matrix.}
	\label{fig:2}
\end{figure}

\begin{figure}[t]
	\centering
	\includegraphics[width=0.9\textwidth]{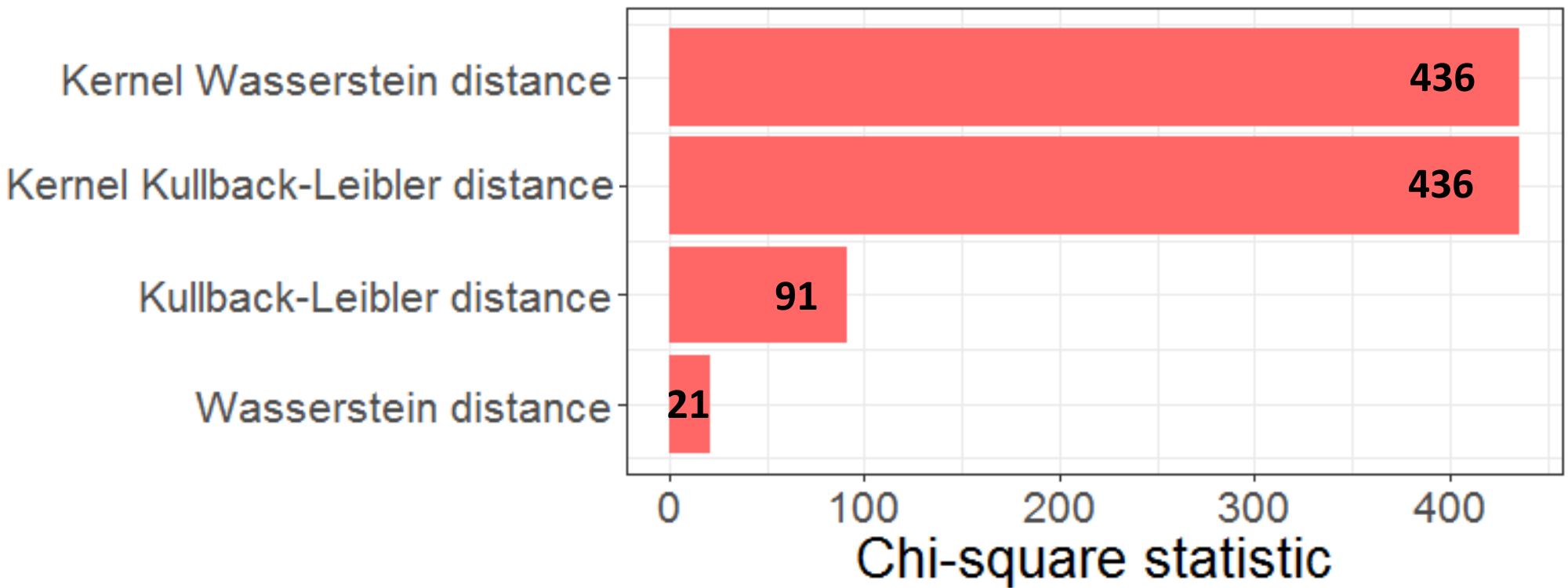}
	\caption{Chi-square statistics for four methods: Wasserstein, kernel Wasserstein, Kullback-Leibler, and kernel Kullback-Leibler distances.}
	\label{fig:3}
\end{figure}

\subsection*{3.2 Texture Features}

Intensity thresholding (at the 5th percentile) was performed to exclude air voxels on each CT slice. The Computational Environment for Radiological Research (CERR) radiomics toolbox was then used to calculate the gray-level co-occurrence matrix (GLCM) from the remaining voxels using 64 gray levels and with a neighborhood of 8 voxels across all 4 directions in 2D \cite{Apte2018,Folkert2017}. A total of 25 scalar features were then extracted from the GLCM (listed in Table 1). For further information on the GLCM features used, see https://github.com/cerr/CERR/wiki/Radiomics.
Each feature was normalized between 0 and 1 for further analysis.

\subsection*{3.3
 Experimental Results}
For 1164 CT slices, we computed kernel Wasserstein distance between each pair of slices  on  25 GLCM-based texture features. After that, we conducted unsupervised hierarchical clustering using the resulting  distance matrix. Figure 2(a) shows a heatmap of the symmetric distance matrix for each pair of slices and Figure 2(b) presents the hierarchical clustering result. We identified two clusters: Cluster 1 with blue lines and Cluster 2 with red lines, consisting of 666 and 498 slices, respectively. In Figure (b), the bar under the hierarchical graph indicates the actual labels with blue for clean slices and  red for noisy slices. As a result, Cluster 1 has 658 clean  and 8 noisy slices whereas Cluster 2 has 230 clean and 268 noisy slices (table in Figure 2(b)). That is, Cluster 1 and 2 were significantly enriched for clean and noisy slices, respectively, with a Chi-square test p-value < 0.0001. Prediction rates were 97.10\% and 74.10\% for noisy and clean slices, respectively, and overall prediction rate was  79.6\%. In Figure 2(a), the  order of slices is the same as that shown in Figure 2(b) and the two clusters were divided by the black dot lines; the left bottom block represents Cluster 1 and the right top block for Cluster 2.  The areas with blue color indicate close  distance between slices whereas the areas with red color indicate far distance. Not surprisingly, the blue areas are mostly shown in the two blocks that represent the distances within each cluster. On the other hand, other two blocks (in the left top and right bottom) mostly have red areas, implying far distance between two clusters (between noisy  and clean slices).

We compared  performance of kernel Wasserstein distance with  other methods, including Wasserstein, KL, and kernel KL distances, using Chi-square statistic. Note that for KL and kernel KL we computed  the average value of two KL measures, i.e., $J_{\mathrm{KL}}(P\|Q)$. We repeated the analysis process noted above for  alternative methods. Of note, kernel methods (kernel Wasserstein distance and kernel KL distance) had the same accuracy with a Chi-square statistic of 436 (Figure 3). By contrast, non-kernel methods (Wasserstein distance and KL distance) had  substantially lower Chi-square statistics  with 21 and 91, respectively, showing the superiority of kernel methods in this application.

Figure 4 shows scatter plots for the correlation between kernel Wasserstein distance and Wasserstein distance in Cluster 1 (Figure 4(a)), Cluster 2 (Figure 4(b)), and between two clusters (Figure 4(c)). Compared to the correlation in Cluster 1 and between two clusters, the correlation in Cluster 2 that was enriched for noisy slices showed relatively more linear pattern. This may suggest that the impact of kernel Wasserstein distance in Cluster 1  and between two clusters is  larger than  classical Wasserstein distance, thereby leading to improved performance.

\begin{figure}
	\centering
	\includegraphics[width=0.9\textwidth]{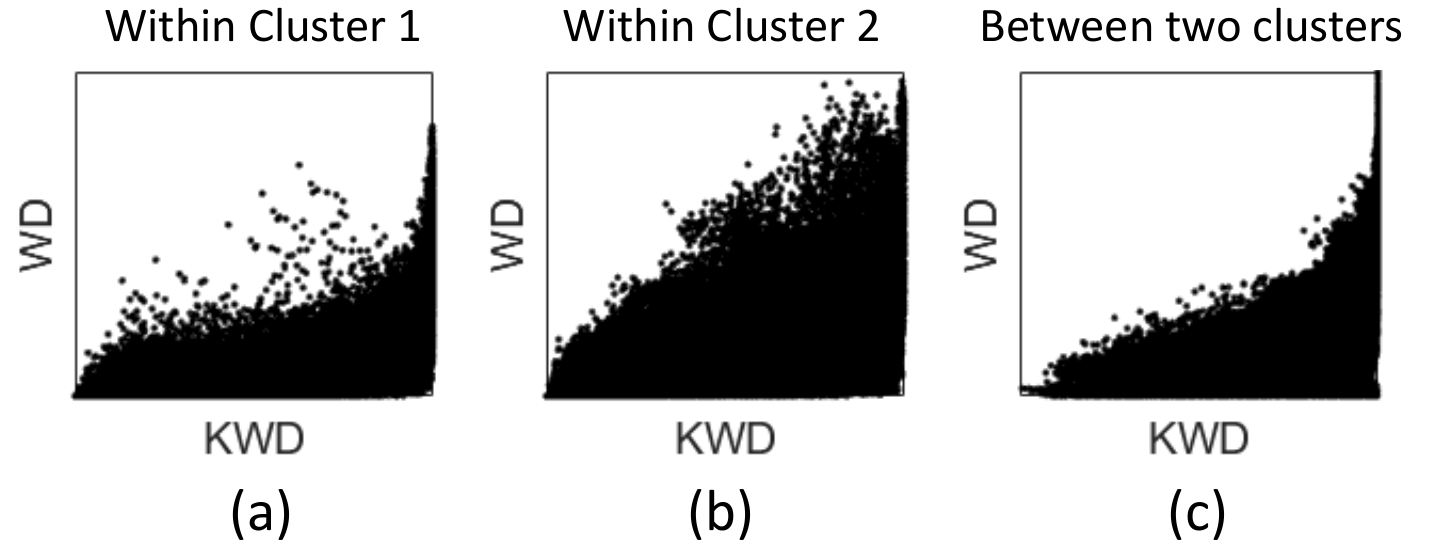}
	\caption{Scatter plots for the correlation between kernel Wasserstein distance and Wasserstein distance in (a) Cluster 1, (b) Cluster 2, and (c) between two clusters. KWD: kernel Wasserstein distance and WD:  Wasserstein distance.}
	\label{fig:4}
\end{figure}

\section*{4 Conclusion}
The Wasserstein distance
is a powerful tool with a wide range of applications. Although extensively used, the method of computing this in the kernel space is lacking. In this paper, we proposed a computational method to solve \textit{L}\textsuperscript{2}-Wasserstein distance in the kernel space. We applied this method to a medical imaging problem in which CT scans are often degraded by artifacts arising from high-density materials. Our unsupervised method consisting of kernel \textit{L}\textsuperscript{2}-Wasserstein distance and hierarchical clustering showed a good level of performance in identifying noisy CT slices, outperforming conventional Wasserstein distance. Notably, kernel Kullback-Leibler distance also obtained comparable performance. This implies the nonlinearity of  data and thus nonlinear analysis using kernel methods would be  more likely to be essential.
Future work will focus on further applications of kernel Wasserstein distance in imaging and biological data analysis.

\subsubsection*{Acknowledgments}
This research was funded in part through National Institutes of Health/National Cancer Institute Cancer Center Support grant P30 CA008748.


\newcommand{\BMCxmlcomment}[1]{}

\BMCxmlcomment{

<refgrp>

<bibl id="B1">
  <title><p>Computational Optimal Transport: With Applications to Data
  Science</p></title>
  <aug>
    <au><snm>Peyre</snm><fnm>G</fnm></au>
    <au><snm>Cuturi</snm><fnm>M</fnm></au>
  </aug>
  <source>Book</source>
  <publisher>Foundations and Trends¡Ëc in Machine Learning</publisher>
  <pubdate>2019</pubdate>
</bibl>

<bibl id="B2">
  <title><p>Pediatric sarcoma data forms a unique cluster measured via the
  earth mover¢®?s distance</p></title>
  <aug>
    <au><snm>Chen</snm><fnm>Y</fnm></au>
    <au><snm>Cruz</snm><fnm>FD</fnm></au>
    <au><snm>Sandhu</snm><fnm>R</fnm></au>
    <au><snm>Kung</snm><fnm>AL</fnm></au>
    <au><snm>Mundi</snm><fnm>P</fnm></au>
    <au><snm>Deasy</snm><fnm>JO</fnm></au>
    <au><snm>Tannenbaum</snm><fnm>A</fnm></au>
  </aug>
  <source>Scientific Reports</source>
  <pubdate>2017</pubdate>
  <volume>7</volume>
  <issue>1</issue>
  <fpage>7035</fpage>
</bibl>

<bibl id="B3">
  <title><p>Optimal transport for Gaussian mixture models</p></title>
  <aug>
    <au><snm>Chen</snm><fnm>Y</fnm></au>
    <au><snm>Georgiou</snm><fnm>TT</fnm></au>
    <au><snm>Tannenbaum</snm><fnm>A</fnm></au>
  </aug>
  <source>IEEE Access</source>
  <pubdate>2019</pubdate>
  <volume>7</volume>
  <fpage>6269</fpage>
  <lpage>6278</lpage>
</bibl>

<bibl id="B4">
  <title><p>Differential properties of sinkhorn approximation for learning with
  wasserstein distance</p></title>
  <aug>
    <au><snm>Luise</snm><fnm>G</fnm></au>
    <au><snm>Rudi</snm><fnm>A</fnm></au>
    <au><snm>Pontil</snm><fnm>M</fnm></au>
    <au><snm>Ciliberto</snm><fnm>C</fnm></au>
  </aug>
  <source>Advances in Neural Information Processing Systems</source>
  <pubdate>2018</pubdate>
  <fpage>5864</fpage>
  <lpage>5874</lpage>
</bibl>

<bibl id="B5">
  <title><p>Area-preservation mapping using optimal mass transport</p></title>
  <aug>
    <au><snm>Zhao</snm><fnm>X</fnm></au>
    <au><snm>Su</snm><fnm>Z</fnm></au>
    <au><snm>Gu</snm><fnm>X.D.</fnm></au>
    <au><snm>Kaufman</snm><fnm>A</fnm></au>
    <au><snm>Sun</snm><fnm>J</fnm></au>
    <au><snm>Gao</snm><fnm>J</fnm></au>
    <au><snm>Luo</snm><fnm>F</fnm></au>
  </aug>
  <source>IEEE Trans Vis Comput Graph.</source>
  <pubdate>2013</pubdate>
  <volume>19</volume>
  <issue>12</issue>
  <fpage>2838</fpage>
  <lpage>2847</lpage>
</bibl>

<bibl id="B6">
  <title><p>Partial Differential Equations and Monge-Kantorovich Mass
  Transfer</p></title>
  <aug>
    <au><snm>Evans</snm><fnm>LC</fnm></au>
  </aug>
  <source>Current Developments in Mathematics</source>
  <pubdate>1999</pubdate>
  <volume>1997</volume>
  <fpage>65</fpage>
  <lpage>126</lpage>
</bibl>

<bibl id="B7">
  <title><p>Topics in optimal transportation</p></title>
  <aug>
    <au><snm>Villani</snm><fnm>C</fnm></au>
  </aug>
  <source>Book</source>
  <publisher>American Mathematical Soc.</publisher>
  <pubdate>2003</pubdate>
</bibl>

<bibl id="B8">
  <title><p>On the translocation of masses, Dokl. Akad. Nauk SSSR 37 (1942)
  227-229, English translation:</p></title>
  <aug>
    <au><snm>Kantorovich</snm><fnm>L.V.</fnm></au>
  </aug>
  <source>Journal of Mathematical Sciences</source>
  <pubdate>2006</pubdate>
  <volume>133</volume>
  <fpage>1381</fpage>
  <lpage>1382</lpage>
</bibl>

<bibl id="B9">
  <title><p>A Novel Integrative Multiomics Method Reveals a Hypoxia-Related
  Subgroup of Breast Cancer with Significantly Decreased Survival</p></title>
  <aug>
    <au><snm>Pouryahya</snm><fnm>M</fnm></au>
    <au><snm>Oh</snm><fnm>JH</fnm></au>
    <au><snm>Javanmard</snm><fnm>P</fnm></au>
    <au><snm>Mathews</snm><fnm>J</fnm></au>
    <au><snm>Belkhatir</snm><fnm>Z</fnm></au>
    <au><snm>Deasy</snm><fnm>J</fnm></au>
    <au><snm>Tannenbaum</snm><fnm>AR</fnm></au>
  </aug>
  <source>bioRxiv</source>
  <pubdate>2019</pubdate>
</bibl>

<bibl id="B10">
  <title><p>Learning from uncertain curves: The 2-Wasserstein metric for
  Gaussian processes</p></title>
  <aug>
    <au><snm>Mallasto</snm><fnm>A</fnm></au>
    <au><snm>Feragen</snm><fnm>A</fnm></au>
  </aug>
  <source>Advances in Neural Information Processing Systems</source>
  <pubdate>2017</pubdate>
  <fpage>5660</fpage>
  <lpage>5670</lpage>
</bibl>

<bibl id="B11">
  <title><p>Generalized Discriminant Analysis Using a Kernel
  Approach</p></title>
  <aug>
    <au><snm>BAUDAT</snm><fnm>G</fnm></au>
    <au><snm>ANOUAR</snm><fnm>F</fnm></au>
  </aug>
  <source>Neural Computation</source>
  <pubdate>2000</pubdate>
  <volume>12</volume>
  <issue>10</issue>
  <fpage>2385</fpage>
  <lpage>2404</lpage>
</bibl>

<bibl id="B12">
  <title><p>Fast kernel discriminant analysis for classification of liver
  cancer mass spectra</p></title>
  <aug>
    <au><snm>Oh</snm><fnm>JH</fnm></au>
    <au><snm>Gao</snm><fnm>J</fnm></au>
  </aug>
  <source>IEEE/ACM Trans Comput Biol Bioinform.</source>
  <pubdate>2009</pubdate>
  <volume>8</volume>
  <issue>6</issue>
  <fpage>1522</fpage>
  <lpage>1534</lpage>
</bibl>

<bibl id="B13">
  <title><p>The kernel trick for distances</p></title>
  <aug>
    <au><snm>Scholkopf</snm><fnm>B.</fnm></au>
  </aug>
  <source>Advances in Neural Information Processing Systems</source>
  <pubdate>2000</pubdate>
  <fpage>301</fpage>
  <lpage>307</lpage>
</bibl>

<bibl id="B14">
  <title><p>Random Features for Large-Scale Kernel Machines</p></title>
  <aug>
    <au><snm>Rahimi</snm><fnm>A</fnm></au>
    <au><snm>Recht</snm><fnm>B</fnm></au>
  </aug>
  <source>Advances in Neural Information Processing Systems</source>
  <pubdate>2007</pubdate>
  <fpage>1177</fpage>
  <lpage>1184</lpage>
</bibl>

<bibl id="B15">
  <title><p>A kernel-based approach for detecting outliers of high-dimensional
  biological data</p></title>
  <aug>
    <au><snm>Oh</snm><fnm>JH</fnm></au>
    <au><snm>Gao</snm><fnm>J</fnm></au>
  </aug>
  <source>BMC Bioinformatics</source>
  <pubdate>2009</pubdate>
  <volume>10(Suppl 4)</volume>
  <fpage>S7</fpage>
</bibl>

<bibl id="B16">
  <title><p>Procrustes Metrics on Covariance Operators and Optimal
  Transportation of Gaussian Processes</p></title>
  <aug>
    <au><snm>Masarotto</snm><fnm>V</fnm></au>
    <au><snm>Panaretos</snm><fnm>VM</fnm></au>
    <au><snm>Zemel</snm><fnm>Y</fnm></au>
  </aug>
  <source>Sankhya A</source>
  <pubdate>2018</pubdate>
  <fpage>1</fpage>
  <lpage>42</lpage>
</bibl>

<bibl id="B17">
  <title><p>The Frechet distance between multivariate normal
  distributions</p></title>
  <aug>
    <au><snm>Dowson</snm><fnm>DC</fnm></au>
    <au><snm>Landau</snm><fnm>B</fnm></au>
  </aug>
  <source>Journal of Multivariate Analysis</source>
  <pubdate>1982</pubdate>
  <volume>12</volume>
  <issue>3</issue>
  <fpage>450</fpage>
  <lpage>455</lpage>
</bibl>

<bibl id="B18">
  <title><p>The distance between two random vectors with given dispersion
  matrices</p></title>
  <aug>
    <au><snm>Olkin</snm><fnm>I</fnm></au>
    <au><snm>Pukelsheim</snm><fnm>F</fnm></au>
  </aug>
  <source>Linear Algebra and its Applications</source>
  <pubdate>1982</pubdate>
  <volume>48</volume>
  <fpage>257</fpage>
  <lpage>263</lpage>
</bibl>

<bibl id="B19">
  <title><p>Wasserstein Riemannian geometry of Gaussian densities</p></title>
  <aug>
    <au><snm>Malago</snm><fnm>L</fnm></au>
    <au><snm>Montrucchio</snm><fnm>L</fnm></au>
    <au><snm>Pistone</snm><fnm>G</fnm></au>
  </aug>
  <source>Information Geometry</source>
  <pubdate>2018</pubdate>
  <volume>1</volume>
  <fpage>137</fpage>
  <lpage>179</lpage>
</bibl>

<bibl id="B20">
  <title><p>The Jordan cononical form of a product of a Hermitian and a
  positive semidefinite matrix</p></title>
  <aug>
    <au><snm>Hong</snm><fnm>Y</fnm></au>
    <au><snm>Horn</snm><fnm>RA</fnm></au>
  </aug>
  <source>Linear Algebra and its Applications</source>
  <pubdate>1991</pubdate>
  <volume>147</volume>
  <fpage>373</fpage>
  <lpage>386</lpage>
</bibl>

<bibl id="B21">
  <title><p>The Multi-Dimensional Hardy Uncertainty Principle and its
  Interpretation in Terms of the Wigner Distribution; Relation With the Notion
  of Symplectic Capacity</p></title>
  <aug>
    <au><snm>Gosson</snm><fnm>M</fnm></au>
    <au><snm>Luef</snm><fnm>F</fnm></au>
  </aug>
  <source>arXiv</source>
  <pubdate>2008</pubdate>
</bibl>

<bibl id="B22">
  <title><p>A Kernel Two-Sample Test</p></title>
  <aug>
    <au><snm>Gretton</snm><fnm>A</fnm></au>
    <au><snm>Borgwardt</snm><fnm>KM</fnm></au>
    <au><snm>Rasch</snm><fnm>MJ</fnm></au>
    <au><snm>Scholkopf</snm><fnm>B</fnm></au>
    <au><snm>Smola</snm><fnm>A</fnm></au>
  </aug>
  <source>Journal of Machine Learning Research</source>
  <pubdate>2012</pubdate>
  <volume>13</volume>
  <fpage>723</fpage>
  <lpage>773</lpage>
</bibl>

<bibl id="B23">
  <title><p>Computing the Kullback-Leibler Divergence between two Weibull
  Distributions</p></title>
  <aug>
    <au><snm>Bauckhage</snm><fnm>C</fnm></au>
  </aug>
  <source>arXiv</source>
  <pubdate>2013</pubdate>
</bibl>

<bibl id="B24">
  <title><p>Using uncorrelated discriminant analysis for tissue classification
  with gene expression data</p></title>
  <aug>
    <au><snm>Ye</snm><fnm>J</fnm></au>
    <au><snm>Li</snm><fnm>T</fnm></au>
    <au><snm>Xiong</snm><fnm>T</fnm></au>
    <au><snm>Janardan</snm><fnm>R</fnm></au>
  </aug>
  <source>IEEE/ACM Trans Comput Biol Bioinform.</source>
  <pubdate>2004</pubdate>
  <volume>1</volume>
  <issue>4</issue>
  <fpage>181</fpage>
  <lpage>190</lpage>
</bibl>

<bibl id="B25">
  <title><p>Robust and accurate cancer classification with gene expression
  profiling</p></title>
  <aug>
    <au><snm>Li</snm><fnm>H</fnm></au>
    <au><snm>Zhang</snm><fnm>K</fnm></au>
    <au><snm>Jiang</snm><fnm>T</fnm></au>
  </aug>
  <source>IEEE Comput Syst Bioinform Conf.</source>
  <pubdate>2005</pubdate>
  <fpage>310</fpage>
  <lpage>321</lpage>
</bibl>

<bibl id="B26">
  <title><p>An optimization criterion for generalized discriminant analysis on
  undersampled problems</p></title>
  <aug>
    <au><snm>Ye</snm><fnm>J</fnm></au>
    <au><snm>Janardan</snm><fnm>R</fnm></au>
    <au><snm>Park</snm><fnm>C.H.</fnm></au>
    <au><snm>Park</snm><fnm>H</fnm></au>
  </aug>
  <source>IEEE Trans. Pattern Anal. Mach. Intell.</source>
  <pubdate>2004</pubdate>
  <volume>26</volume>
  <issue>8</issue>
  <fpage>982</fpage>
  <lpage>994</lpage>
</bibl>

<bibl id="B27">
  <title><p>Sherman-morrison-woodbury-formula-based algorithms for the surface
  smoothing problem</p></title>
  <aug>
    <au><snm>Lai</snm><fnm>S.H.</fnm></au>
    <au><snm>Vemuri</snm><fnm>B.C.</fnm></au>
  </aug>
  <source>Linear Algebra and its Applications</source>
  <pubdate>1997</pubdate>
  <volume>265</volume>
  <fpage>203</fpage>
  <lpage>229</lpage>
</bibl>

<bibl id="B28">
  <title><p>Computational Medicine in Data Mining and Modeling</p></title>
  <aug>
    <au><snm>Rakocevic</snm><fnm>G</fnm></au>
    <au><snm>Djukic</snm><fnm>T</fnm></au>
    <au><snm>Filipovic</snm><fnm>N</fnm></au>
    <au><snm>Milutinovic</snm><fnm>V</fnm></au>
  </aug>
  <source>Book</source>
  <publisher>Springer New York</publisher>
  <pubdate>2013</pubdate>
</bibl>

<bibl id="B29">
  <title><p>Technical Note: Extension of CERR for computational radiomics: A
  comprehensive MATLAB platform for reproducible radiomics research</p></title>
  <aug>
    <au><snm>Apte</snm><fnm>AP</fnm></au>
    <au><snm>Iyer</snm><fnm>A</fnm></au>
    <au><snm>Crispin Ortuzar</snm><fnm>M</fnm></au>
    <au><snm>Pandya</snm><fnm>R</fnm></au>
    <au><snm>Dijk</snm><fnm>LV</fnm></au>
    <au><snm>Spezi</snm><fnm>E</fnm></au>
    <au><snm>Thor</snm><fnm>M</fnm></au>
    <au><snm>Um</snm><fnm>H</fnm></au>
    <au><snm>Veeraraghavan</snm><fnm>H</fnm></au>
    <au><snm>Oh</snm><fnm>JH</fnm></au>
    <au><snm>Shukla Dave</snm><fnm>A</fnm></au>
    <au><snm>Deasy</snm><fnm>JO</fnm></au>
  </aug>
  <source>Med. Phys.</source>
  <pubdate>2018</pubdate>
  <volume>45</volume>
  <issue>8</issue>
  <fpage>3713</fpage>
  <lpage>3720</lpage>
</bibl>

<bibl id="B30">
  <title><p>Predictive modeling of outcomes following definitive
  chemoradiotherapy for oropharyngeal cancer based on FDG-PET image
  characteristics</p></title>
  <aug>
    <au><snm>Folkert</snm><fnm>MR</fnm></au>
    <au><snm>Setton</snm><fnm>J</fnm></au>
    <au><snm>Apte</snm><fnm>AP</fnm></au>
    <au><snm>Grkovski</snm><fnm>M</fnm></au>
    <au><snm>Young</snm><fnm>RJ</fnm></au>
    <au><snm>Schoder</snm><fnm>H</fnm></au>
    <au><snm>Thorstad</snm><fnm>WL</fnm></au>
    <au><snm>Lee</snm><fnm>NY</fnm></au>
    <au><snm>Deasy</snm><fnm>JO</fnm></au>
    <au><snm>Oh</snm><fnm>JH</fnm></au>
  </aug>
  <source>Phys Med Biol.</source>
  <pubdate>2017</pubdate>
  <volume>62</volume>
  <issue>13</issue>
  <fpage>5327</fpage>
  <lpage>5343</lpage>
</bibl>

</refgrp>
} 

\medskip

\small

\end{document}